\documentclass{article}
\usepackage[utf8]{inputenc}
\usepackage{amsfonts}
\usepackage{amsmath}
\usepackage[percent]{overpic}
\usepackage{verbatim}
\usepackage{xcolor}
\usepackage{mdframed}
\usepackage{xstring}
\usepackage{multirow}
\usepackage{anyfontsize}
\usepackage{tabularx}
\usepackage{wrapfig}
\usepackage{datatool}
\DTLsetseparator{:}
\usepackage{xifthen}
\usepackage{caption}
\usepackage{subcaption}
\usepackage{multirow}

\usepackage{listofitems}
\usepackage{floatrow}
\usepackage[affil-it]{authblk}
\usepackage{ifthen}
\usepackage{comment}
\usepackage{bbm}
\usepackage[many]{tcolorbox}
\usepackage[euler]{textgreek}
\usepackage{algorithm}
\usepackage{textcomp}
\usepackage[noend]{algpseudocode}
\usepackage{url}

\makeatletter
\def\BState{\State\hskip-\ALG@thistlm}
\makeatother

\usepackage{transparent}
\usepackage[margin=1in]{geometry}
\title{A Machine Learning Outlook: Post-processing of Global Medium-range Forecasts}
\author[1]{Shreya Agrawal}
\author[1]{Rob Carver}
\author[1]{Cenk Gazen}
\author[2]{Eric Maddy}
\author[3]{Vladimir Krasnopolsky}
\author[1]{Carla Bromberg}
\author[1]{Zack Ontiveros}
\author[1]{Tyler Russell}
\author[1]{Jason Hickey}
\author[3]{Sid Boukabara}

\affil[1]{Research, Google}
\affil[2]{Riverside Technology, Inc. at NOAA}
\affil[3]{National Oceanic and Atmospheric Administration}

\date{\vspace{-5ex}}
\usepackage{natbib}
\usepackage{graphicx}

\begin{document}

\maketitle

\begin{abstract}
    Post-processing typically takes the outputs of a Numerical Weather Prediction (NWP) model and applies linear statistical techniques to produce improve localized forecasts, by including additional observations, or determining systematic errors at a finer scale. In this pilot study, we investigate the benefits and challenges of using non-linear neural network (NN) based methods to post-process multiple weather features -- temperature, moisture, wind, geopotential height, precipitable water -- at 30 vertical levels, globally and at lead times up to 7 days. We show that we can achieve accuracy improvements of up to 12\% (RMSE) in a field such as temperature at 850hPa for a 7 day forecast. However, we recognize the need to strengthen foundational work on objectively measuring a sharp and correct forecast. We discuss the challenges of using standard metrics such as root mean squared error (RMSE) or anomaly correlation coefficient (ACC) as we move from linear statistical models to more complex non-linear machine learning approaches for post-processing global weather forecasts.
\end{abstract}

\section{Review of previous efforts}
\subsection{Traditional methods for post-processing}
The first approach that was used for statistical post-processing was known as Model Output Statistics, \citet{Glahn1972-sy}, based on multiple multilinear regressions. The U.S. National Weather Service has used these methods to improve systematic model errors since 1968, \citet{Carter1989-gi, Wilks2007-fz}. This approach has also been applied to correct errors in ensembles becoming Ensemble Model Output Statistics (EMOS), \citet{Gneiting2005-uu}. These methods demonstrate significant reduction of errors in numerical forecasts, \citet{Hemri2014-mu}. However, there are several significant limitations:
\begin{itemize}
    \item They are essentially linear techniques. To account for the nonlinear character of errors (e.g., due to different atmospheric regimes, terrain types, etc.), multiple multilinear regressions are introduced to correct errors in different variables, at different locations, and under different weather conditions. This tremendously increases the number of linear regressions used by the system.
    \item They require a significant amount of additional information about statistical properties of parameters, \citet{Haupt2021-er}.
\end{itemize}

\subsection{Can Machine Learning perform better?}
Recently, several machine learning techniques have shown promising results compared to traditional numerical methods based on physics equations such as \citet{Dueben2018-kl}, \citet{Espeholt2022-xe}, \citet{Weyn2021-jl}, and more specifically for post-processing in both probabilistic and deterministic settings (\citet{Taillardat2016-st, Campos2020-ze, Gronquist2021-am, Cho2022-rz, Kirkwood2021-je}). \citet{Krasnopolsky2012-or} show that a shallow NN applied to post-process a set of multi-model ensembles for 24 hour precipitation forecasts, over continental US, show small improvements over taking an arithmetic mean of the ensemble members in terms of regular statistical scores. However, it significantly improves sharpness of meteorological features and reduces the number of false alarms. \citet{Rasp2018-ec} use NNs for probabilistic post-processing of ensemble members predicting 2m temperature over Germany and report the continuous ranked probability scores. They use feature importance techniques to gain insight into the non-linear relationships between the predictors and predictands. \citet{Chapman2019-dn} post-process NCEP’s Global Forecast system forecast of integrated vapor transport over parts of the US, on a 0.5 degree grid, and report significant RMSE and ACC improvements.

\section{Our Approach}
Here, we present our approach and results on post-processing a total of 173 weather features over the entire globe at a 0.25 degree resolution, from 6 hours to 7 days out. We use NOAA’s Global forecast system (GFS) as the physics-based model to post-process and the global data assimilation system (GDAS) as the target of our model. Our goal from using one model to post-process all the weather features together was to allow the model to learn all kinds of inter-dependencies between variables. Note that, GFS itself is a constituent of GDAS. Hence, one of our other objectives is to use the output of the bias corrected model as an input into other data assimilation systems.
\subsection{Training and Validation Details}
We use data from all of 2019 and 2020 for training and 2021 for completely independent testing. This allows us to evaluate robustly whether this method can be applied to future forecasts. This meant that our training procedure used two different versions of GFS (v14 (pre July 2019, spectral dycore) and v15 (July 2019-March 2021, FV3 dycore)) and our testing was majorly on a different version (v15 and v16 (post March 2021, FV3 dycore)). Table \ref{table:1} describes the weather variables we consider for post-processing.

\begin{table}[h!]
\begin{center}
\begin{tabular}{ |p{5cm}|p{5cm}|p{5cm}| } 
\hline
Temperature & \multirow{4}{5cm}{10hPa - 1000hPa, every 50hPa} \\ 
Geopotential Height & \\ 
U and V-component of Wind & \\ 
Relative Humidity & \\ 
\hline
Cloud Mixing Ratio & 10hPa - 1000hPa, every 50hPa \\
\hline
Precipitable Water & \multirow{3}{5cm}{Single level variables} \\ 
Surface Pressure & \\ 
Total Ozone & \\ 
\hline
\end{tabular}
\end{center}
\caption{Summary of the post-processed weather features.}
\label{table:1}
\end{table}

We now describe the training variables used as inputs and targets. Suppose GFS makes a 6hr forecast at current time t, denoted by $GFS_t^6$, that we want to post-process and bring closer to the actual observation at t+6, denoted by $GDAS_{t+6}$. Let’s define the error in the physics-based model forecast, $e$, as 
\begin{equation}
    e_{t+6}=GFS_t^6-GDAS_{t+6}
\end{equation} . The target of our prediction is therefore $e_{t+6}$. The inputs to ML model are the following:
\begin{itemize}
\item Latitude, longitude, azimuth and altitude of the sun for the target time t+6.
\item $GFS_t^6$- The forecast to be post-processed.
\item$GDAS_{t}$- The analysis or state of the world at t.
\item $B_N$- The weighted average of historical errors.
\item The weighted average of historical errors, $B_N$, is given by
\begin{equation}
B_{N} = \frac{\sum_{i=0}^{N}e_{t-i}(1-w^t)}{\sum_{i=0}^N\left(1-w^t\right)}
\end{equation}where $w^t$ is a constant and we use ($1-w^t)$ to decay the weight as we consider historical error further away in time.
\end{itemize}

We use a technique adapted from \citet{Cui2012-pb} to compute a statistical bias using the decaying weighted average of historical errors as an input. In the ideal scenario, one would instead feed the historical errors directly into the model instead of computing an average first. However, due to memory constraints that is challenging when we are trying to post-process 173 features together over the entire globe.

The global data is projected to the equirectangular projection at $0.25X0.25$ degree resolution. This results in a $1440X721$ field for each variable, further multiplied by 173 variables to post-process. To fit all of the input training data into memory, we use a tiling method to split the global field into multiple tiles of $256X256$ grid points each. 

\subsection{Model Architecture considerations}
We considered a few different neural network architectures to post-process. One was the well-known UNet architecture with residual skip connections and the other was a pointwise network with three layers of convolutions with filter size 64, 128 and 256 respectively and a $1X1$ kernel size. We believed that a simpler network would demonstrate whether neighboring pixels have any influence on increasing the accuracy of any given pixel for post-processing. As it turns out, the simpler model with ~576K parameters, performed as well as the more complex UNet architecture with ~82M parameters. Therefore, here we present results for the simpler model with similar accuracy. We address this model as \textit{ML-PP} from here on.

We considered three choices for training the tiled samples (1) use all of the tiles from across the globe into a single model that weighs all grid points at any lat-lon equally, (2) weigh the loss function by the cosine of the latitude, (3) train three separate models, for the northern extratropics, tropical belt, and southern extratropics. The first choice of training performed poorly compared to the other two, due to the equirectangular projection overly weighting the grid points around the poles. Using a weighted loss function performed better since we could now compensate for the poor projection, however still, the error e  made by the GFS models is much larger in the extra-tropics than in the tropical belt. This minimizes the error reduction in the tropical belt. The third choice of training performed the best, allowing the models to learn differently over the extra-tropics than the tropical belt.

For comparison, we trained our model on several standard loss functions such as mean squared error (MSE), mean absolute error (MAE) and Log Cosh--the latter two were included because they are less sensitive to outlying incorrect predictions. We also trained on some functions to closely resemble metrics used in meteorology: the cosine similarity loss function that mimics correlation coefficient, and fractions skill score that effectively computes MSE at several different spatial resolutions. We found that the results of training over these loss functions, in aggregate over the entire globe, were not statistically significantly different from each other and hence, we report on results using MSE as the loss function. We discuss later the need for another kind of loss function and metric.

\section{Results}
Here we present our results and compare them with two baseline models:

\textbf{Linear model:} This model is intended to mimic the linear regressions of Model Output Statistics. This model is highly localized.  It takes a series of historical model errors and the VSDB climatology, \citet{Hoffman2017-ec}, of expected values at a grid point and makes a forecast correction for that grid point only. 

\textbf{Weighted Decay Average:} This uses the statistically computed bias $B_N$ and subtracts that from the forecast, $GFS_t^6$, to obtain the corrected forecast (\citet{Cui2012-pb}).

\textbf{Gaussian Blur:} We post-process GFS by applying a spatial Gaussian blur to it with a sigma of 2. This helps us understand how well the ML model performs as compared to a simple blurring of the forecast.

\begin{figure}[ht]
\begin{center}
   \centering
  \includegraphics[width=1\linewidth]{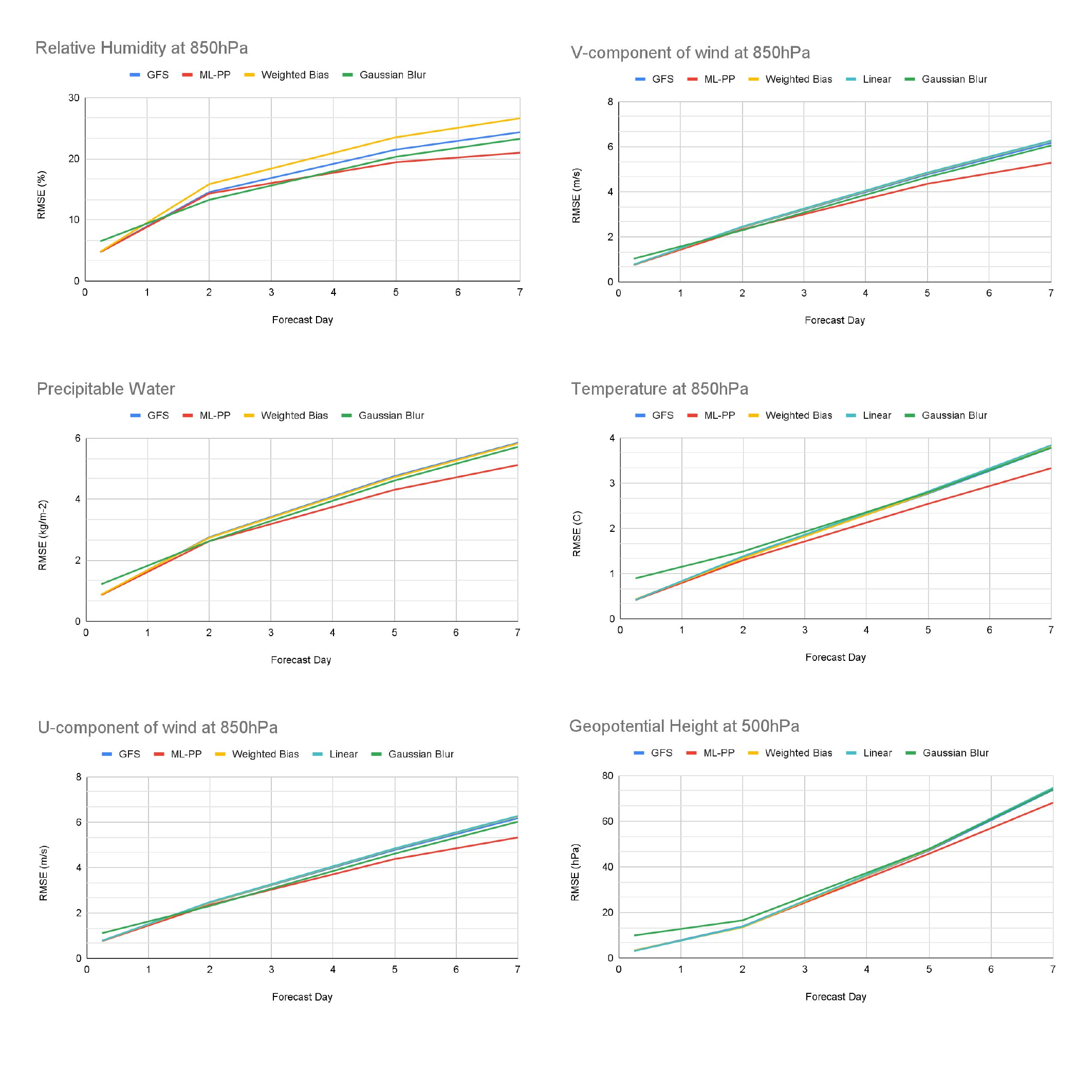}
\caption{RMSE of different weather variables at 500hPa and 850hPa.}
\label{fig:fig_1}
\end{center}
\end{figure}

We present results for a few key weather variables that we think are of interest to the meteorological community at large. In Fig. \ref{fig:fig_1}, we show results for root mean squared errors. We see improvements of $\approx$7.6\% for geopotential height at 500hPa, $\approx$12\% for temperature and $\approx$14\% for the wind components, at 850hPa, when using \textit{ML-PP} for post-processing up to 7 days out. This is a significant improvement over GFS. For the surface variables, we see improvements in accuracy as well such as $\approx$12.5\% for precipitable water, $\approx$13.8\% for relative humidity and $\approx$7\% for surface pressure.

\begin{figure}[ht]
\begin{center}
   \centering
  \includegraphics[width=1\linewidth]{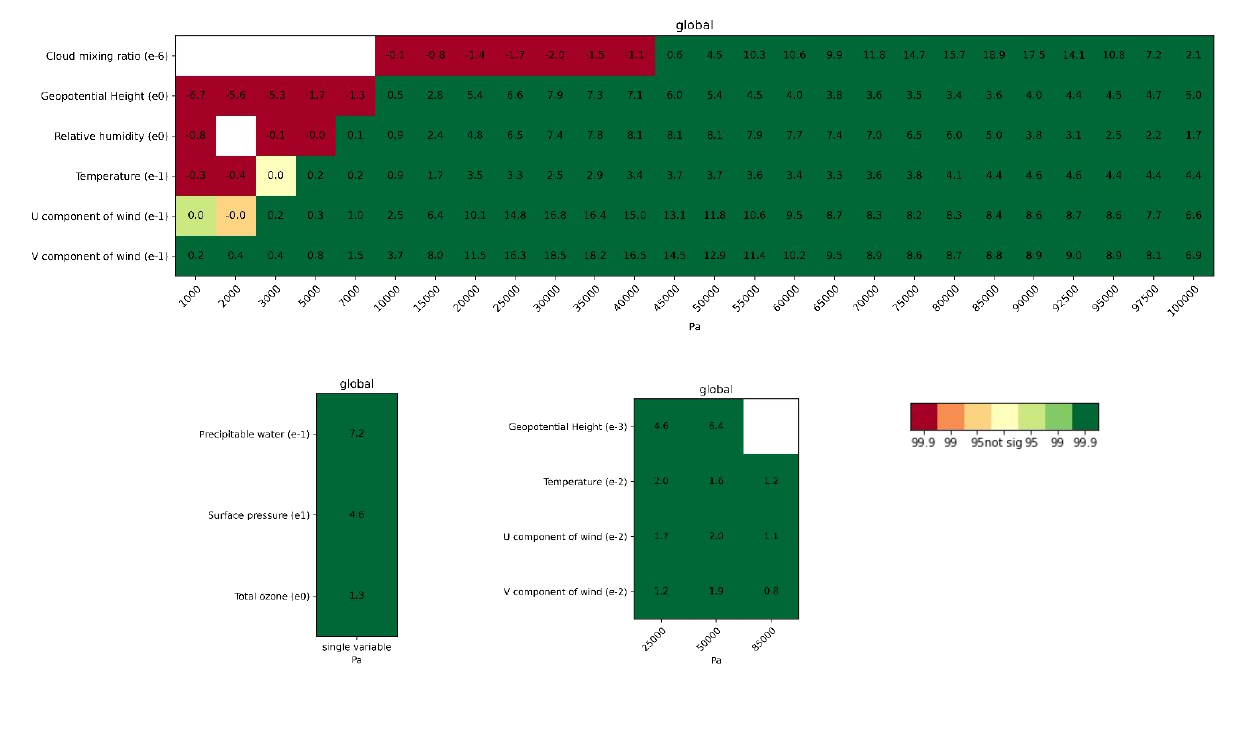}
\caption{Skill card with color-coding to indicate statistical significance for 7 day post-processing at different vertical levels. White blanks mean data at those levels was not available in GFS. Top: MSE for variables at multiple vertical levels, bottom-left: MSE for surface variables, bottom-right: Anomaly Correlation Coefficient, “not sig” in the center of colorbar means not significant.}
\label{fig:fig_2}
\end{center}
\end{figure}

In Fig. \ref{fig:fig_2}, we summarize the differences in RMSE and Anomaly Correlation Coefficient between post-processing by \textit{ML-PP} and GFS in color-coded tables. The color coding indicates whether the difference is statistically significant and at what level. We compute statistical significance using the two-tailed t-distribution looking at the population of differences across all time steps. Green colors indicate that post-processing improves the metrics (i.e., RMSE is smaller and Anomaly Correlation Coefficient is larger after post-processing). The fields with single values are surface and total column fields. For RMSE, there is a marginal improvement across most of the variables, and for Anomaly Correlation Coefficient, we see small improvements for temperature and the U-V components of wind but none for geopotential height. 

\section{Re-thinking Evaluation Metrics}
\begin{figure}
     \centering
     \begin{subfigure}{\textwidth}
         \centering
         \includegraphics[width=\textwidth]{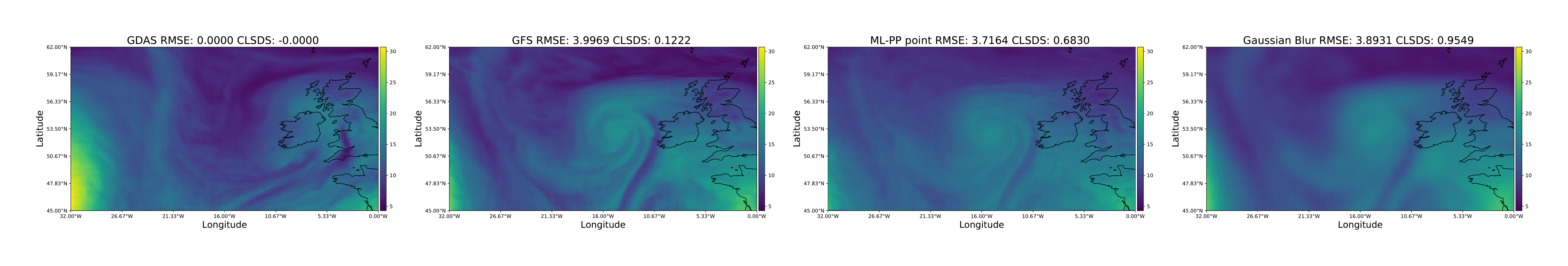}
         \caption{Forecast of Precipitable Water. Left: GDAS ground truth, then GFS forecast, then ML post-processed forecast (ML-PP), right-most: gaussian blurred GFS forecast.}
         \label{fig:y equals x}
     \end{subfigure}
     \hfill
     \begin{subfigure}[b]{\textwidth}
         \centering
         \includegraphics[width=\textwidth]{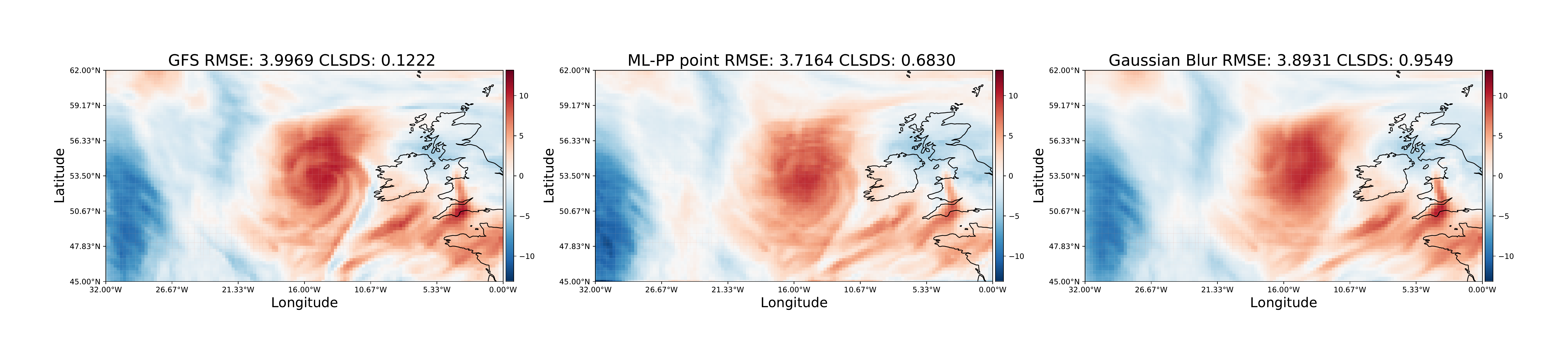}
         \caption{Model Error of each forecast from GDAS for Precipitable water. Left: GFS forecast error, middle: ML post-processed forecast (ML-PP) error, right-most: gaussian blurred GFS forecast error.}
         \label{fig:three sin x}
     \end{subfigure}
     \hfill
     \begin{subfigure}[b]{\textwidth}
         \centering
         \includegraphics[width=\textwidth]{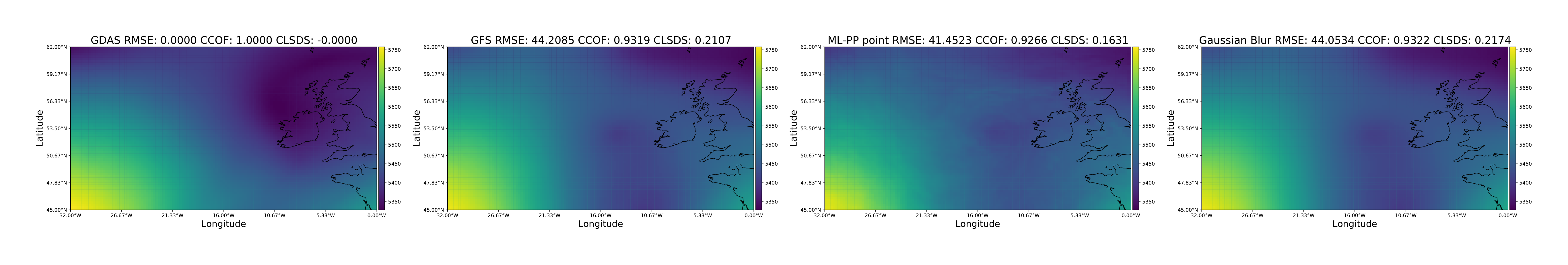}
         \caption{Forecast of Geopotential Height at 500hPa showing super-resolution. Left: GDAS ground truth, then GFS forecast, then ML post-processed forecast (ML-PP), right-most: gaussian blurred GFS forecast.}
         \label{fig:five over x}
     \end{subfigure}
     \hfill
     \begin{subfigure}[b]{\textwidth}
         \centering
         \includegraphics[width=\textwidth]{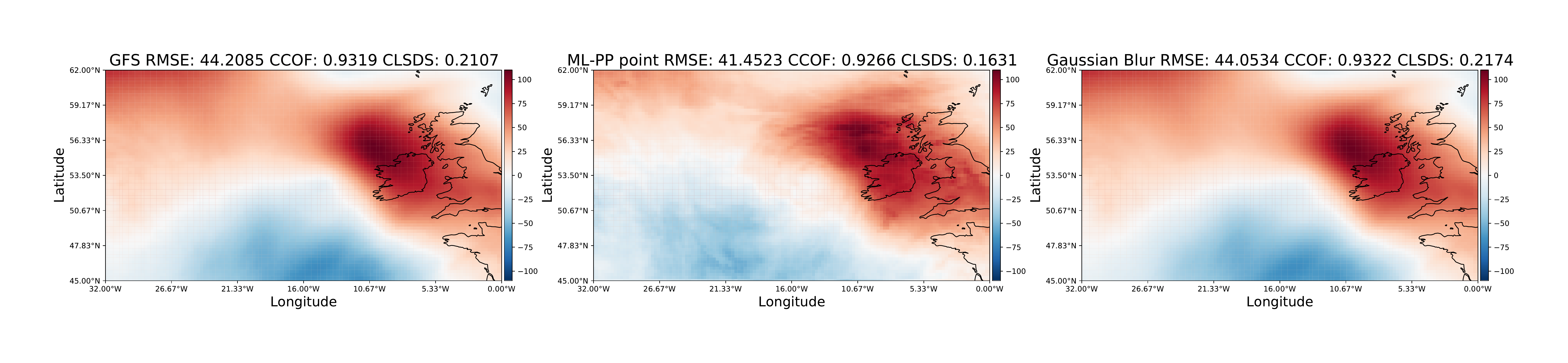}
         \caption{Model Error of each forecast from GDAS for Geopotential Height at 500hPa. Left: GFS forecast error, middle: ML post-processed forecast (ML-PP) error, right-most: gaussian blurred GFS forecast error.}
         \label{fig:3d}
     \end{subfigure}
\caption{Case study of Storm Darcy (5th February, 2021 00hrs UTC): Comparison of 5 day forecast made by GFS, ML post-processed forecast, and forecast obtained through gaussian blurring of GFS.}
\label{fig:fig_3}
\end{figure}

In the previous section, we saw that for some standard metrics that are common to ML methods and meteorology, such as MSE and ACC, the neural network model performs reasonably well. However, we examined a few case studies and noticed small amounts of blurriness in the predictions. This blurriness is inherent in convolutional neural networks trained to predict the mean of the spread. In Fig. \ref{fig:fig_3}, we evaluate Storm Darcy that occurred at the beginning of February 2021, and brought high winds, heavy snowfall and cold temperatures to Ireland, UK, Germany and Netherlands. Here, when predicting a 5 day forecast for 5th February, 2021, Model \textit{ML-PP} is able to reduce RMSE by ~0.28 kg/m2  and \textit{Gaussian Blur} reduces it by ~0.1 kg/m2. However, looking at the forecast, it is surprising that the \textit{Gaussian Blur} model, which is intentionally blurred, manages to lower MSE.

To capture this notion of blurriness introduced by the post-processing methods, we use a calibrated log spectral distance skill score (CLSDS), similar to the one used by \citet{Ravuri2021-qa}. CLSDS is a measure of (potentially spurious) high spatial frequency structure in the geophysical fields produced by the various models P relative to base O. The skill score is a normalized mean of differences between the log spectrums of input fields. We compute the spectral distance, $S_d$ with:

\begin{equation}
    S_{d}(P,O) = \overline{\left(\frac{\log({S(P))}}{B(P)} - \frac{\log(S(O))}{B(O)}\right)}
\end{equation}

Where $S$, spectrum, is computed via FFT and $B$ is the bias component in the spectrum. The absolute values computed this way are hard to interpret so we normalize the distance by dividing it by the distance  between the base field and a Gaussian-blurred version of the base field. It is positive if P is blurrier than base, negative if it is sharper. It is normalized so that a CLSDS of 1 is equivalent to applying $G_{1}$, a Gaussian filter with sigma 1 (default) to $O$. 
\begin{equation}
    \textrm{CLSDS} = \frac{S_{d}(P,O)}{S_{d}(G_{1}(P), O)}
\end{equation}

The normalization depends only on the base so clsds(P, O) != clsds(O, P). This is an experiment to figure out ways to measure blurriness introduced by ML models.

\begin{figure}[ht]
\begin{center}
   \centering
  \includegraphics[width=1\linewidth]{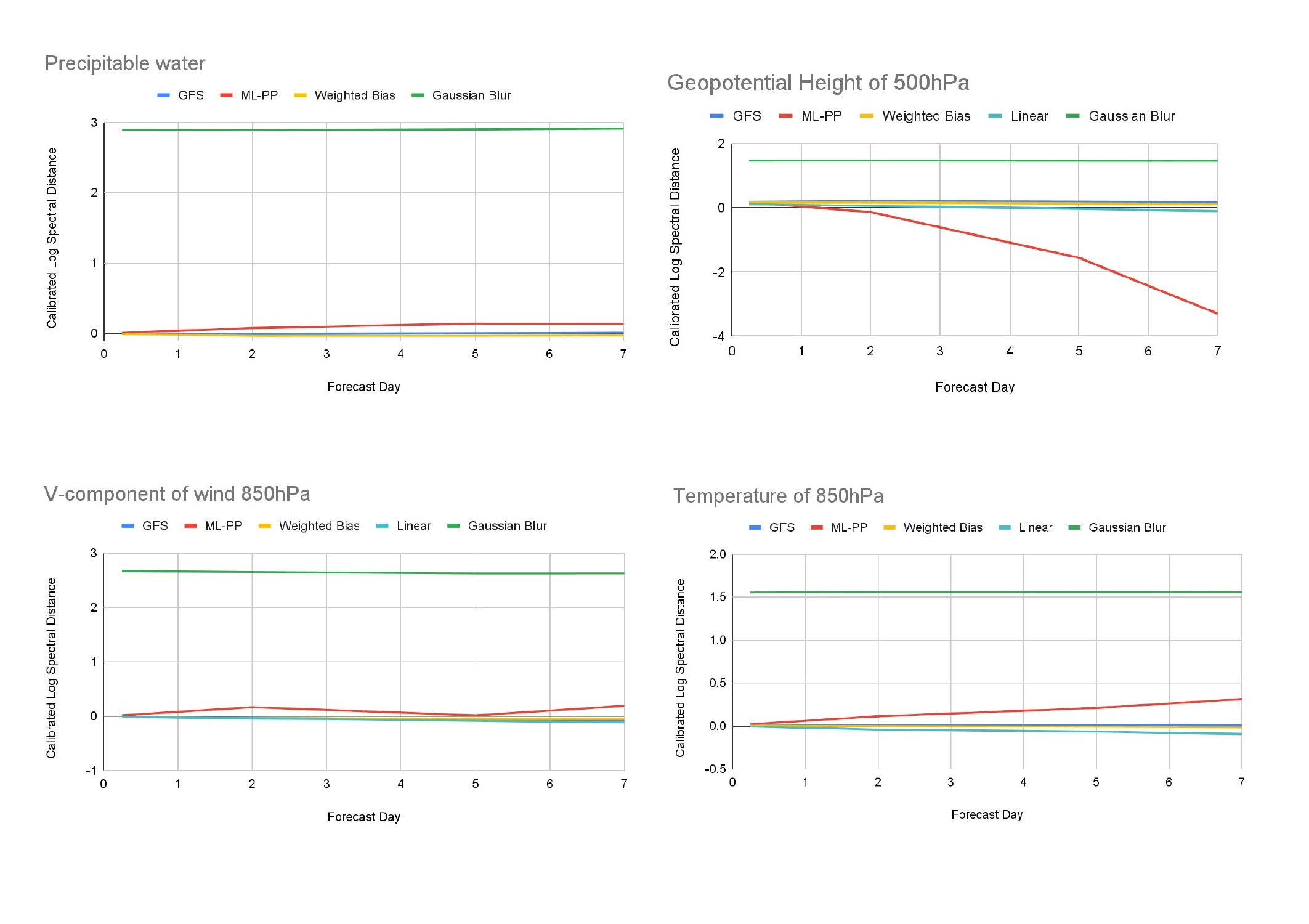}
\caption{Calibrated Log Spectral Distance Skill for various lead times.}
\label{fig:fig_4}
\end{center}
\end{figure}

When using CLSDS metric on our models, GDAS becomes the base and we evaluate GFS, \textit{ML-PP} and \textit{Gaussian blur} against it. In the case study above, \textit{Gaussian blur} gets a significantly high score of 0.882, while \textit{ML-PP} gets 0.577 which is still blurrier than GFS at 0.026. In Fig. \ref{fig:fig_4}, that shows the results for various lead times, note that while the \textit{ML-PP} model is blurrier than GFS for all variables except geopotential height, \textit{Gaussian blur} is much more blurry. For geopotential height, it is interesting to note that post-processing seems to have decreased the blurriness of GFS, by super-resolving it.

\section{Conclusion}
\citet{Lorenz1982-gp} posited that a desirable characteristic of forecast maps produced by NWP models is that they should look like real weather maps.  We wish to extend this guideline to forecast maps produced by post-processing NWP output.  In this pilot study, we have shown that ML-based post-processing schemes can improve performance on forecast verification metrics such as MSE and ACC.  When visualized, the post-processed fields are smoother than the NWP input and the eventual verification analysis.  

Our results suggest a paradigm shift is necessary for selecting loss functions in training ML post-processing schemes.  A loss function based on MSE will train a model that truncates extremes and diffuses sharp horizontal gradients. This model will minimize MSE and boost ACC, but these gains come at the expense of an unnaturally smooth forecast field.  We call on the meteorological/machine learning community to address these challenges in two ways:
\begin{itemize}
    \item Instead of predicting a smoothed ensemble mean, we can post-process individual ensemble members and create new estimates of ensemble spread and mean from the post-processed ensemble.
    \item Investigate loss metrics that will minimize error while preserving the scale and magnitude of synoptic variability. Addressing these challenges will be essential for building a trustworthy post-processing system.  
\end{itemize}

\section{Acknowledgements}
The authors would like to thank Joseph Mani, Kevin Garrett, Fei Sha and Phatty Arbuckle for their valuable contribution in organizing this work. We would also like to thank Sanjay Agravat for his engineering contributions and Stephan Hoyer for his valuable reviews.

\nocite{*}

\bibliographystyle{unsrtnat}

\bibliography{references}
\end{document}